\documentclass[sigconf]{acmart}
\AtBeginDocument{%
  }

\copyrightyear{2026}
\acmYear{2026}
\setcopyright{cc}
\setcctype{by} 
\acmConference[WWW '26] {Proceedings of the ACM Web Conference 2026}{April 13--17, 2026}{Dubai, United Arab Emirates}
\acmBooktitle{Proceedings of the ACM Web Conference 2026 (WWW '26), April 13--17, 2026, Dubai, United Arab Emirates}
\acmISBN{979-8-4007-2307-0/2026/04}
\acmDOI{10.1145/3774904.3792891}

\usepackage{balance} 
\usepackage{graphicx}  
\usepackage{multirow}  
\usepackage{booktabs}  
\usepackage{enumitem}

\begin{document}

\title{CytoCrowd: A Multi-Annotator Benchmark Dataset for Cytology Image Analysis}


\author{Yonghao Si}
\email{siyh3@mail2.sysu.edu.cn}
\affiliation{%
  \institution{Sun Yat-sen University}
  \institution{Hong Kong University of Science and Technology (Guang Zhou)} 
  \city{Guangzhou}   
  \country{China}
}

\author{Xingyuan Zeng}
\email{zengxy96@mail2.sysu.edu.cn}
\affiliation{%
  \institution{Sun Yat-sen University}
  \city{Guangzhou}
  \country{China}
}

\author{Zhao Chen}
\email{chenzhao@hkust-gz.edu.cn}
\affiliation{%
  \institution{Hong Kong University of Science and Technology (Guang Zhou)}
  \city{Guangzhou}
  \country{China}
}

\author{Libin Zheng}
\authornote{Corresponding author.}
\email{zhenglb6@mail.sysu.edu.cn}
\affiliation{%
  \institution{Sun Yat-sen University}
  \city{Guangzhou}
  \country{China}
}

\author{Caleb Chen Cao}
\email{cao@ust.hk}

\author{Lei Chen}
\email{leichen@cse.ust.hk}
\affiliation{%
  \institution{Hong Kong University of Science and Technology}
  \city{Hong Kong}
  \country{China}
}

\author{Jian Yin}
\email{issjyin@mail.sysu.edu.cn}
\affiliation{%
  \institution{Sun Yat-sen University} 
  \city{Guangzhou}
  \country{China}
}

\renewcommand{\shortauthors}{Yonghao Si et al.}

\begin{abstract}
    High-quality annotated datasets are crucial for advancing machine learning in medical image analysis. However, a critical gap exists: most datasets either offer a single, clean ground truth, which hides real-world expert disagreement, or they provide multiple annotations without a separate gold standard for objective evaluation. To bridge this gap, we introduce CytoCrowd, a new public benchmark for cytology analysis. The dataset features 446 high-resolution images, each with two key components: (1) raw, conflicting annotations from four independent pathologists, and (2) a separate, high-quality gold-standard ground truth established by a senior expert. This dual structure makes CytoCrowd a versatile resource. It serves as a benchmark for standard computer vision tasks, such as object detection and classification, using the ground truth. Simultaneously, it provides a realistic testbed for evaluating annotation aggregation algorithms that must resolve expert disagreements. We provide comprehensive baseline results for both tasks. Our experiments demonstrate the challenges presented by CytoCrowd and establish its value as a resource for developing the next generation of models for medical image analysis. The dataset is publicly available at \url{https://github.com/YHSI5358/CytoCrowd-Datasets}.
\end{abstract}



\maketitle

\section{Introduction}

Machine learning models are now important tools in medical image analysis. They can assist doctors with diagnoses and expedite research. However, the success of these models depends heavily on large, high-quality annotated datasets.
The need for accurate data is especially high in complex fields, such as cytology. Cytology images are difficult to annotate because they contain many overlapping cells with small but important differences. As a result, even expert pathologists often disagree on the exact boundaries, categories, or even the existence of certain objects.

Existing datasets do not fully address this challenge. General-purpose datasets like COCO~\cite{DBLP:conf/eccv/LinMBHPRDZ14} are not suitable for medical tasks, as they lack the specific complexities of cytology images. Many specialized medical datasets, such as BraTS~\cite{menze2014multimodal}, solve this by providing a single, unified ground truth. This approach is useful for basic model training, but it hides the real-world disagreements that occur between experts. This makes it difficult to develop models that can handle uncertainty.

Other datasets, like LIDC-IDRI~\cite{samuel2011lung}, do provide annotations from multiple experts. This is useful for studying expert disagreement. However, they typically lack a separate gold-standard ground truth that has been verified by a senior expert. Without a final, trusted reference, it is difficult to reliably evaluate and compare different algorithms. As shown in Table~\ref{tab:dataset_comparison}, a benchmark is needed that offers \textbf{both} the raw expert disagreements and a separate, high-quality gold-standard ground truth.

\begin{table*}[ht]
    \centering
    \begin{tabular}{lcccc}
    \toprule
    \textbf{Dataset} & \textbf{Domain} & \textbf{Raw Disagreements} & \textbf{Gold-Standard GT} & \textbf{Primary Use Case}  \\
    \midrule
    COCO~\cite{DBLP:conf/eccv/LinMBHPRDZ14} & General Objects & No & Yes & CV Models  \\
    BraTS~\cite{menze2014multimodal} & Brain Tumors & No & Yes (Consensus) & CV Segmentation  \\
    LIDC-IDRI~\cite{samuel2011lung} & Lung Nodules & Yes & No (Consensus serves as GT) & Nodule Analysis  \\
    \textbf{CytoCrowd (Ours)} & \textbf{Cytology} & \textbf{Yes (4 physicians)} & \textbf{Yes (Senior expert)} & \textbf{CV \& Aggregation Benchmark} \\
    \bottomrule
    \end{tabular}
    \caption{A comparison of CytoCrowd with other datasets. CytoCrowd is unique because it provides both raw expert disagreements and a separate gold-standard ground truth.}
    \label{tab:dataset_comparison}
    \vspace{-20pt}
\end{table*}

To fill this gap, we introduce \textbf{CytoCrowd},\footnote{The complete dataset is publicly available at \url{https://github.com/YHSI5358/CytoCrowd-Datasets}.} a new public dataset for cytology analysis. CytoCrowd is built to support two main research areas. The dataset contains 446 high-resolution images. Four independent pathologists annotated these images, creating 14,579 raw annotations that show real-world clinical disagreements. Importantly, a senior pathologist then reviewed all annotations to create a final gold-standard ground truth of 6,402 objects for reliable evaluation.

This structure makes CytoCrowd a flexible resource:
\begin{itemize}
    \item For \textbf{Computer Vision} researchers, the gold-standard ground truth provides a clear benchmark for training and testing models for object detection, segmentation, and classification.
    \item For \textbf{Crowdsourcing} researchers, the raw expert annotations serve as a realistic testbed for algorithms that aim to combine multiple, conflicting annotations into a single, accurate result.
\end{itemize}

In summary, our contributions are:
\begin{enumerate}
    \item A new public dataset with both raw expert annotations and a separate gold-standard ground truth for cytology.
    \item A benchmark that captures real-world expert disagreements on object boundaries, categories, and existence.
    \item Baseline results for both computer vision and annotation aggregation methods to support future work.
\end{enumerate}

\section{Related Work}

Our work is related to two main categories of datasets: standard benchmarks for medical computer vision, with a focus on cytology, and datasets containing annotations from multiple experts.

\noindent\textbf{Datasets for Medical Computer Vision.} Deep learning models require large, annotated datasets for training. Many high-quality medical imaging datasets have been created for this purpose. For example, the BraTS dataset~\cite{menze2014multimodal} provides images for brain tumor segmentation, while the KiTS dataset~\cite{heller2019kits19} focuses on kidney tumor segmentation. These datasets have been essential for advancing the field.

A common characteristic of these benchmarks is that they typically provide a single, clean ground truth for each image. This ground truth is often created by having several experts create annotations, which are then merged into a single, final version. This approach is practical and provides a clear target for training standard segmentation or detection models.

However, this method of providing a single, pre-consolidated ground truth has a limitation. It hides the natural ambiguities and disagreements that are common in clinical practice. In reality, experts may disagree on the precise edges of a region or on its diagnostic classification. By removing this information, these datasets make it difficult to develop or evaluate models that can reason about uncertainty or are robust to variations in annotation style.

\noindent\textbf{Multi-Annotator Datasets in Cytology.} To address the issue of expert disagreement, some datasets provide raw annotations from multiple experts. The LIDC-IDRI dataset~\cite{samuel2011lung} is a well-known example, containing lung nodule segmentations from four radiologists. Another example is VinDr-CXR~\cite{nguyen2022vindr}, which includes labels for chest X-rays from multiple annotators.
These datasets are extremely valuable. They allow researchers to study the extent of inter-observer variability and provide a realistic basis for developing annotation aggregation algorithms. These algorithms aim to intelligently combine multiple, potentially conflicting labels into a single, more reliable result.

While these datasets are crucial for aggregation research, they often have a limitation when it comes to evaluation. They typically lack a separate, definitive gold-standard ground truth that stands apart from the initial annotators' opinions. For instance, the ``true" label might be defined as the consensus of the annotators. This makes it challenging to perform a truly objective evaluation, as there is no independent reference to compare against.

Our work, CytoCrowd, is designed to fill the gaps left by both categories of datasets. It provides the raw, conflicting annotations necessary for aggregation research, while also offering the separate, gold-standard ground truth needed for clear and objective evaluation of computer vision models.

\section{The CytoCrowd Dataset}

\begin{figure}[t]
  \centering
  \includegraphics[width=\linewidth]{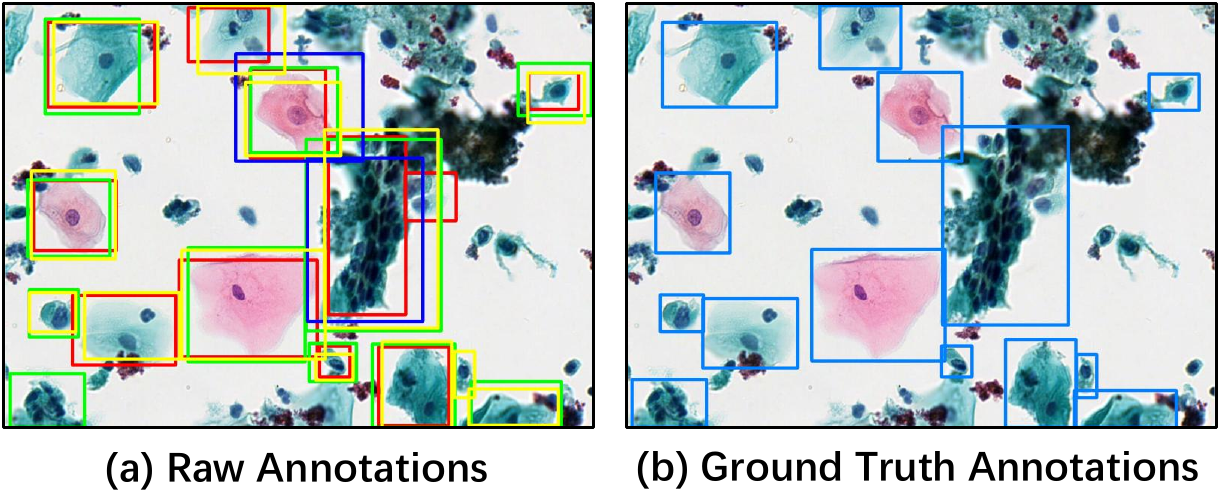} 
  \caption{Raw expert annotations (left) vs. the final gold-standard ground truth (right) on a sample image from CytoCrowd.}
  \label{fig:challenges}
  \vspace{-20pt}
\end{figure}

The CytoCrowd dataset is the result of a seven-month collaborative project between the Hong Kong University of Science and Technology (Guangzhou)\footnote{\url{https://www.hkust-gz.edu.cn/}} and Guangzhou LBP Medicine Science \& Technology Co.\footnote{\url{https://en.gzlbp.com/}}. It contains 446 high-resolution cytology images, which are acquired using a whole-slide scanner at 40x magnification and stored in .svs format. This dataset is specifically created to provide a public benchmark for developing and evaluating annotation aggregation algorithms in the context of complex medical images. The complete dataset, including the raw pathologist annotations and the senior-verified gold-standard ground truth, is publicly available at \url{https://github.com/YHSI5358/CytoCrowd-Datasets}.

The annotation process is conducted by four board-certified pathologists from LBP, each possessing over ten years of clinical experience. To ensure the independence of each expert's opinion, they are asked to annotate the images separately. Using a custom-developed annotation platform\footnote{\url{https://mdi.hkust-gz.edu.cn/metal/user/}} from the Hong Kong University of Science and Technology (Guangzhou), each pathologist identifies objects of interest by drawing bounding boxes (Regions of Interest, ROIs) and assigning a diagnostic category from a predefined set of 34 classes. This independent process generated a total of 14,579 raw annotations, capturing a wide range of inter-observer disagreements on object boundaries, categories, and existence. Figure~\ref{fig:challenges} visually illustrates this core feature of our dataset. The left panel shows the raw annotations for a sample region, where different colors highlight the conflicting labels from different physicians. The right panel shows the single, consolidated gold-standard ground truth for the same region, which serves as the definitive reference for evaluation.

To create a definitive ground truth for evaluation, a senior pathologist with more than fifteen years of experience meticulously reviewed every one of the 14,579 annotations. This expert's task is to consolidate, correct, and finalize the set of ROIs and their corresponding categories for each image, establishing a gold-standard reference. This rigorous process resulted in a final ground truth of 6,402 objects. 


\paragraph{Expert Disagreement Analysis.}
The complexity of the dataset is further evidenced by the inter-annotator variability. The average pairwise Intersection over Union (IoU) is 0.664, reflecting localized disagreement. Crucially, regarding object existence, only 11.37\% of cells are identified by all four experts, whereas 34.78\% are annotated by a single expert. This creates a high ratio of raw annotations to ground truth objects (approx. 2.28 to 1), underscoring the necessity of our senior-verified gold standard as a reliable benchmark. The key statistics of the CytoCrowd dataset are summarized in Table~\ref{tab:stats}.

\begin{table}[h!]
\centering

\begin{tabular}{@{}lr@{}}
\toprule
\textbf{Statistic} & \textbf{Value} \\ \midrule
\# Workers (Physicians) & 4 \\
\# Tasks (Images) & 446 \\
\# Total Annotations (ROIs) & 14,579 \\
\# Ground Truth Objects (GT ROIs) & 6,402 \\
\# Categories & 34 \\ \bottomrule
\end{tabular}
\caption{Statistics of the CytoCrowd Dataset.}
\label{tab:stats}
\vspace{-30pt}
\end{table}
\section{Experiments and Baselines}

\subsection{Task Definition}
\label{sec:task_definition}

The CytoCrowd benchmark is designed to evaluate methods for two distinct tasks, each targeting a different research community.

\noindent\textbf{Task 1: Medical Object Detection and Classification}

This is a standard computer vision task. The goal is to train a model that can directly identify the location and diagnostic category of cellular objects in the images.
\begin{itemize}[left=0pt]
    \item \textbf{Input:} The high-resolution cytology images.
    \item \textbf{Training Data:} Models are trained using the \textbf{gold-standard ground truth} annotations.
    \item \textbf{Goal:} For a given test image, the model should produce a set of bounding boxes and a corresponding class label for each box.
    \item \textbf{Evaluation:} The model's predictions are compared against the gold-standard ground truth of the test set.
\end{itemize}

\noindent\textbf{Task 2: Annotation Aggregation}

This task is for the crowdsourcing and truth inference community. The goal is to combine the conflicting annotations from multiple experts into a single, high-quality result that is as close to the ground truth as possible.
\begin{itemize}[left=0pt]
    \item \textbf{Input:} The raw, and often conflicting, annotations from the four independent pathologists for a given image.
    \item \textbf{Goal:} The algorithm should process multiple inputs and produce a single, consolidated set of annotations (bounding boxes and class labels).
    \item \textbf{Evaluation:} The algorithm's consolidated output is compared against the \textbf{gold-standard ground truth}.
\end{itemize}

By defining these two separate tasks, we provide a clear benchmark for both learning-based vision models and aggregation algorithms.

\subsection{Evaluation Metrics}

To evaluate performance, we report the classification \textbf{Accuracy}. This metric is calculated in a way that separates localization success from classification correctness. Specifically, we first match predicted objects to ground truth objects using the standard \textbf{IoU} metric. A prediction is considered correctly located---a \textbf{true positive (TP)}---if its IoU with a ground truth object is greater than 0.5. The final Accuracy is then calculated as the percentage of correctly classified objects \textit{only among these TPs}. This approach ensures that the accuracy score purely reflects the model's ability to classify objects that it has successfully located, which provides a clear and fair comparison across different methods.

\subsection{Baseline Methods}
\label{sec:baselines}

We provide baseline results for the two tasks defined for the CytoCrowd dataset. These baselines are chosen to represent common or powerful approaches in their respective fields.

\subsubsection{Annotation Aggregation Baselines}

For the annotation aggregation task, we evaluate well-known inference methods including \textbf{Majority Voting (MV)}, the statistical \textbf{Dawid \& Skene (D\&S)~\cite{dawid1979maximum}} model, and truth discovery methods such as \textbf{CATD~\cite{DBLP:journals/pvldb/LiLGSZDFH14}}, \textbf{PM~\cite{DBLP:conf/sigmod/LiLGZFH14}}, \textbf{LFC~\cite{DBLP:journals/jmlr/RaykarYZVFBM10}}, and \textbf{ZenCrowd~\cite{DBLP:conf/www/DemartiniDC12}}.

\subsubsection{Learning-based Baselines}
For the medical object detection and classification task, we test several modern, powerful vision models instead of traditional object detectors. 
\begin{itemize}[left=0pt]
    \item \textbf{DeepEdit~\cite{DBLP:journals/corr/abs-2305-10655}} and \textbf{Anytime \cite{DBLP:journals/corr/abs-2403-15218}:} These are models based on or related to interactive and prompt-based segmentation, representing the state-of-the-art in producing precise object masks.
    \item \textbf{Qwen-VL-MAX~\cite{DBLP:journals/corr/abs-2308-12966}} and \textbf{Qwen2.5-VL-72B~\cite{DBLP:journals/corr/abs-2502-13923}:} These are large-scale Vision-Language Models (VLMs). We test them to see if their extensive general-world knowledge can be applied to this specialized medical task.
\end{itemize}

\subsection{Performance Analysis}

We present the performance of the baseline methods on the two defined tasks.

\subsubsection{Performance of Annotation Aggregation Methods}
Table~\ref{tab:category_results} shows the accuracy results for the annotation aggregation baselines. The performance is measured by comparing the aggregated category labels against the gold-standard ground truth for all correctly located objects.

The results are very informative. Interestingly, the simplest baseline, \textbf{Majority Voting (MV), achieves the highest accuracy (0.903)}, outperforming more complex models like Dawid \& Skene (D\&S). This finding is significant as it suggests that when all annotators are domain experts, their collective agreement provides a powerful signal. In this scenario, complex models that try to learn and correct for annotator errors may not provide an advantage, as the experts are all highly reliable. This result highlights the unique nature of our expert-annotated dataset and presents a challenge to existing aggregation methods.

\begin{table}[t]
    \centering
    \small
    
    \begin{tabular}{lc}
        \toprule
        \textbf{Method}  & Accuracy \\
        \midrule
        CATD & 0.857  \\
        Dawid \& Skene (D\&S) & 0.893 \\
        Majority Voting (MV) & 0.903 \\
        PM & 0.855  \\
        LFC  & 0.896  \\
        ZenCrowd  & 0.883  \\
        \bottomrule
    \end{tabular}
    
    \caption{\textbf{Performance of annotation aggregation methods. Accuracy is calculated on correctly localized objects (TPs).}}
    \label{tab:category_results}
    \vspace{-20pt}
\end{table}

\subsubsection{Performance of Learning-based Methods}
Table~\ref{tab:cv_result} presents the accuracy of the vision models on our dataset. The results show a clear and consistent trend.

There is a major performance gap between the general-purpose VLMs and the more specialized segmentation models. The \textbf{Qwen VLMs perform very poorly}, with accuracy below 45\%. This strongly indicates that despite their vast knowledge, these large models cannot effectively handle the fine-grained and domain-specific challenges of cytology image analysis without specialized fine-tuning.

In contrast, the models more focused on segmentation, \textbf{DeepEdit and Anytime, achieve high accuracy (0.899 and 0.878, respectively)}. Their strong performance establishes a solid and competitive baseline for future computer vision research. This demonstrates the CytoCrowd dataset's value as a benchmark for developing and testing new, specialized models that address challenges missed by general models.

\begin{table}[t]
    \centering
    \small
    
    \begin{tabular}{lc}
        \toprule
        \textbf{Method} & Accuracy \\
        \midrule
        Qwen-VL-MAX & 0.441\\
        Qwen2.5-VL-72B & 0.437 \\
        DeepEdit  &  0.899  \\
        Anytime  & 0.878  \\
        \bottomrule
    \end{tabular}
    
    \caption{\textbf{Comparison on CV/VLM baselines}.}
    \label{tab:cv_result}
    \vspace{-20pt}
\end{table}

\section{Conclusion}
We have introduced \textbf{CytoCrowd}, a new expert-annotated benchmark for complex medical image annotation aggregation. Our extensive benchmarking demonstrates the difficulty of the task. We acknowledge two limitations: the dataset size (446 images) is relatively small compared to general CV benchmarks, and the single-expert gold standard may introduce observer bias. Future iterations could incorporate biopsy-verified labels. Despite this, we hope CytoCrowd will spur the development of novel algorithms that can effectively harness the collective intelligence from multiple, conflicting expert annotations.
\begin{acks}
This work is supported by the National Natural Science Foundation of China (No. 62472455, U22B2060); National Key Research and Development Program of China (Grant No. 2023YFF0725100); Key-Area Research and Development Program of Guangdong Province (2024B0101050005); Guangdong-Hong Kong Technology Innovation Joint Funding Scheme (Project No. 2024A0505040012); Key Areas Special Project of Guangdong Provincial Universities (2024ZDZX1006); Guangdong Province Science and Technology Plan Project (2023A0505030011); Research Foundation of Science and Technology Plan Project of Guangzhou City (2023B01J0001, 2024B01W0004); Guangzhou municipality big data intelligence key lab (2023A03J0012); AOE Project (AoE/E-603/18); Theme-based project TRS (T41-603/20R); CRF Project (C2004-21G); HKUST(GZ) - CMCC(Guangzhou Branch) Metaverse Joint Innovation Lab (Grant No. P00659); Hong Kong ITC ITF grant (PRP/004/22FX); Zhujiang scholar program (2021JC02X170); and HKUST-Webank joint research lab.
\end{acks}

\bibliographystyle{ACM-Reference-Format}
\balance
\bibliography{sample-base}

\end{document}